\documentclass{INTERSPEECH2023}

\interspeechcameraready

\title{N-best T5: Robust ASR Error Correction using Multiple Input Hypotheses and Constrained Decoding Space}
\name{Rao Ma, Mark J. F. Gales, Kate M. Knill, Mengjie Qian \thanks{This paper reports on research supported by Cambridge University Press \& Assessment, a department of The Chancellor, Masters, and Scholars of the University of Cambridge.}}
\address{ALTA Institute, Machine Intelligence Lab, Department of Engineering, Cambridge University, UK}
\email{\{rm2114,mjfg100,kmk1001,mq227\}@cam.ac.uk}

\usepackage{multirow, multicol}
\usepackage{algorithm, algorithmic, tabularx}

\begin{document}
\maketitle
 
\begin{abstract}
Error correction models form an important part of Automatic Speech Recognition (ASR) post-processing to improve the readability and quality of transcriptions. Most prior works use the 1-best ASR hypothesis as input and therefore can only perform correction by leveraging the context within one sentence. In this work, we propose a novel N-best T5 model for this task, which is fine-tuned from a T5 model and utilizes ASR N-best lists as model input. By transferring knowledge from the pre-trained language model and obtaining richer information from the ASR decoding space, the proposed approach outperforms a strong Conformer-Transducer baseline. Another issue with standard error correction is that the generation process is not well-guided. To address this a constrained decoding process, either based on the N-best list or an ASR lattice, is used which allows additional information to be propagated.
% Moreover, we experiment with different decoding algorithms in the generation process. By constraining the decoding space into an N-best list or ASR lattice, further performance improvement can be obtained on the Librispeech test sets.
\end{abstract}

\noindent\textbf{Index Terms}: speech recognition, error correction, ASR N-best list, path merging, lattice-constrained decoding

\section{Introduction}
Automated speech recognition (ASR) refers to the advanced technology of transcribing human speech into readable text. Early ASR systems employed HMM-based architectures where different modules such as the acoustic model, language model, and pronunciation lexicon, were trained separately and combined effectively during the decoding process. Recently, with the development of computing resources and availability of high-quality speech data, end-to-end (E2E) ASR models such as LAS and RNN-T have shown better performance and become more prevailing in both academia and industry \cite{graves2014towards, chan2016listen, gulati2020conformer}. Many successful applications have been built based on large-scale E2E ASR models with the aim of helping targeted users increase their productivity \cite{hoy2018alexa}.

Although ASR systems have shown good performance in general, even beating human transcribers in terms of recognition accuracy in some experimental settings, they still face challenges when deployed in practice \cite{radford2022robust}. 
% For instance, scenarios where multi-people are speaking, the presence of background noise, and when spoken words rarely appear in the training speech corpus can all hinder the performance of ASR models. This model is expected to identify errors contained in the ASR output automatically and correct them based on other words within the sentence. within a successful business system,
Therefore, error correction models that are expected to correct errors within the ASR output continue to serve an important role in post-processing~\cite{errattahi2018automatic}. Recently, end-to-end error correction models have shown promising performance on this task \cite{hrinchuk2020correction, zhang2019automatic, dutta2022error, zhao2021bart}.
Most works on error correction share a similar structure, taking the 1-best recognized hypothesis from the ASR system as input. The reference text is used as the target in training and beam search is used for generation in inference. We argue that this standard practice does not provide enough information to the encoder and also gives the model too much freedom in decoding, which results in degraded performance.

The error correction model does not have access to the original utterance and makes predictions based on the ASR output. When the error correction model uses the 1-best hypothesis as input, it can only make predictions based on words within a single sentence. This sentence, however, contains limited information about the utterance and might be erroneous.
%Consequently, the model is prone to make mistakes with the limited input information.
% Since the correction model does not have access to the original utterance and can only make predictions based on the erroneous ASR output, it is prone to make mistakes. 
Different strategies have been proposed to tackle this problem. Some works focus on obtaining and utilizing a more compact output from the ASR system such as the N-best list \cite{zhu2021improving, leng2021fastcorrect}, a word confusion net (WCN) \cite{weng2020joint}, and a lattice \cite{ma2020neural, dai2022latticebart}.
% rather than the plain 1-best decoding hypothesis. 
Other works propose using additional features such as phoneme sequences \cite{wang2020asr} in the correction process. These modifications enable the error correction model to obtain more useful information, making error detection and correction easier to perform.

To boost the performance of error correction models, we propose to fine-tune from a pre-trained language model (PLM) using concatenated N-best lists as input. With this extended input, the error correction model can effectively compare the difference between each hypothesis and becomes aware of the positions where the ASR system might have made mistakes. Since the oracle WER in the N-best list is relatively low, words within the N-best list also serve as a kind of ``cues'' to help the model recover the correct tokens. By fine-tuning from a high performing text-to-text transfer transformer (T5) model \cite{raffel2020exploring}, we can take advantage of the structured knowledge implicitly learned from wide-ranging text data. As PLMs are pre-trained on large-scale text corpus via unsupervised learning, we hypothesize that these models can be easily adapted to plain text inputs, such as N-best lists, rather than more complex representations such as a lattice or WCN.

% In this work, we choose to use the concatenated N-best list as input rather than using more complicated data structures such as lattice or WCN. 

Error correction models usually use beam search in decoding, which we refer to as unconstrained decoding in this paper. For PLMs, words with similar meanings that share similar contexts are usually close to each other in the embedding parameter space. In some cases the model incorrectly generates 
% words that have similar meanings but sound very different from the original utterance. 
synonyms that sound very different from the original utterance.
To improve the robustness of the error correction model, we assume that it should only output hypotheses that also have a high ASR probability. To achieve this, we generate lattices in the decoding of the Transducer ASR model and propose several constrained decoding algorithms for the correction model.
% With this lattice-constrained decoding, we can help obtain correction output that also has a high ASR probability. 
% In the decoding, we also develop methods to utilize the probability scores generated by the ASR model.

The contributions of our work are as follows: This is the first paper in the error correction area to transfer knowledge from a pre-trained language model using  N-best lists as input. Only a simple data augmentation method, SpecAugment, is needed for training data preparation. Compared to a strong Conformer-Transducer ASR baseline, we achieved a 7\% WER reduction on test sets by using N-best lists as input. To further improve model performance we introduce a novel lattice generation and conversion algorithm to the ASR Transducer model. By constraining the decoding space of the N-best T5 model we achieved up to 12\% WER reduction on the test sets.

\section{Discussion on ASR Output}
The error correction model, as a downstream task of speech recognition, aims to correct errors contained in ASR transcriptions. Therefore, the input of the correction model is dependent on the output of the ASR system. In this section we discuss the different types of ASR outputs utilized in our experiments.

E2E ASR models typically employ the beam search algorithm to find the best decoding result. Suppose the beam size is $n$, at each time step a sentence list containing $n$-best partial hypotheses is maintained and updated. When the decoding process is completed for an utterance, we can get an ASR N-best list in addition to the 1-best recognition hypothesis.

A word lattice represents multiple ASR hypotheses and associated scores in a compact graph structure. It has been extensively exploited in HMM-based ASR systems \cite{xu2018pruned, zhang2019lattice, liu2016two}. Recent works attempt to produce lattices during the decoding of E2E ASR models \cite{prabhavalkar2021less, novak2022rnn}. \cite{prabhavalkar2021less} introduces K-gram context approximation to RNN-T, which assumes that the prediction network only needs to rely on the previous $k-1$ tokens instead of the entire predicted sentence history for effective estimation. Based on this assumption, a path-merging algorithm is proposed for beam search. At each decoding step, two partial hypotheses are merged if their last $k-1$ tokens are identical. Then paths with lower scores are removed from the active beam but are kept in the lattice graph. 

\begin{figure}[!htbp]
\centering
\includegraphics[width=0.4\textwidth]{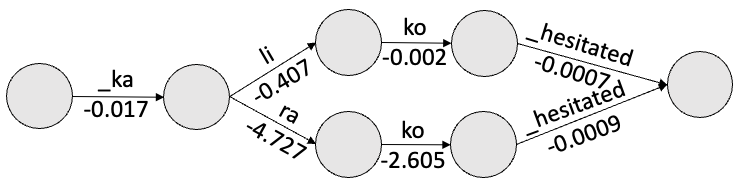}
\caption{An example of the generated ASR lattice.}
\label{asr_bpe}
\vspace{-2mm}
\end{figure}

An example of a generated lattice with $k=3$ is depicted in Figure \ref{asr_bpe}. 
% Merging similar paths in the decoding allows for more diversified hypotheses in the active beam and produces a lattice that might be useful for downstream NLP tasks \cite{prabhavalkar2021less}. 
Past work has shown that the oracle WER in the lattice is reduced compared to the 1-best ASR hypothesis \cite{prabhavalkar2021less}, however, whether the usage of the generated lattice can benefit downstream tasks remains to be demonstrated. In this paper, we show how the lattice produced in the Transducer decoding can help to improve the performance of the error correction system.

\section{N-best T5 Model}

\subsection{Model Structure}
% TODO: Modify picture, text prefix

A traditional error correction model uses the 1-best hypothesis generated by the ASR model as model input. With such limited information, it is difficult for the model to detect the errors or to make the right corrections. Meanwhile, an N-best list is always output by the ASR model as a byproduct of beam search. Hypotheses within this list are highly possible transcriptions of the original utterance and yield lower WER results compared to the 1-best hypothesis. By leveraging the diversified hypotheses contained in the N-best list, the error correction model can make more accurate and informed predictions.

\begin{figure}[hbtp!]
\centering
\includegraphics[width=0.47\textwidth]{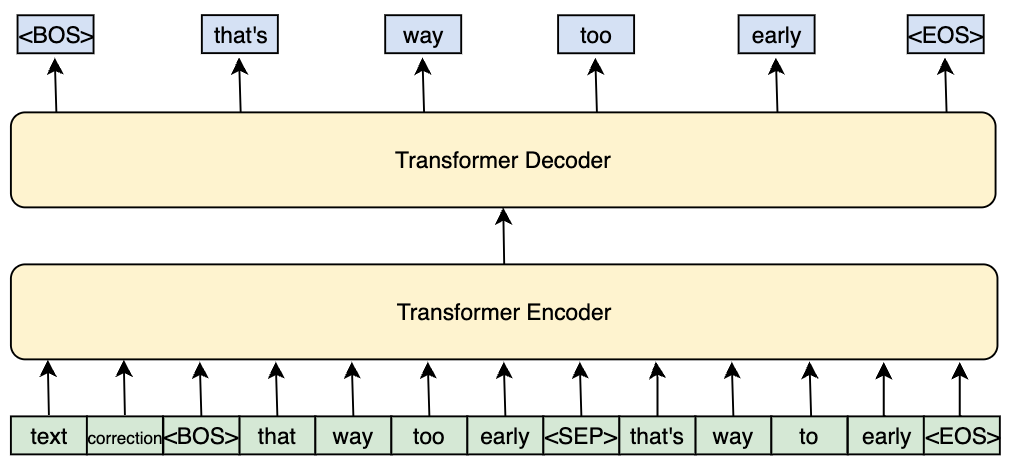}
\caption{Model structure of N-best T5.}
\label{fig:n-best}
\vspace{-3mm}
\end{figure}

The proposed N-best T5 error correction model is shown in Figure~\ref{fig:n-best}. It is based on a pre-trained T5 model with some modifications. N refers to how many top hypotheses we obtain from the ASR system, which is a hyper-parameter to be examined. In training and decoding, we concatenate the ASR N-best list and add a special token between different hypotheses to denote the sentence end. To comply with the pre-training practice of T5, a prefix ``\textit{text correction:}'' is inserted to the input sequence. Preliminary experimental results show that without using the prefix, performance drops on the validation set. 

Since the language model structure remains the same in the fine-tuning phase, the proposed method easily works with any PLM from third-party packages. Furthermore, as the N-best list can be easily obtained from ASR systems without the need of in-depth access, this error correction method can be applied to black-box ASR models integrated in cloud services.
% The reference text is given to the decoder and the model parameters are trained with cross-entropy loss. 

\subsection{Constrained Decoding in Inference}
% \subsubsection{ASR Baseline}
% \begin{equation}
%     \hat{\bm x}_{1:N} = {\arg\max}_{\bm x_{1:N}} \log P(\bm x_{1:N}|\texttt{utt};\bm\theta_\text{ASR})
% \end{equation}
So far we have introduced how to incorporate more information into the encoder part of our proposed model. How to provide guidance for decoding and achieving controllable generation is another interesting question.
The beam search algorithm is widely used as an approximation for finding the best generation result within the entire decoding space.  This gives the model too much freedom, however, and we do not have much control over this decoding process \cite{dathathriplug}. For example, we hope the proposed model keeps correct words from the original transcription and only corrects incorrect words. In addition, the model is expected to output homophones of detected erroneous words. Although the model can implicitly learn these features from the training data, there is no guarantee in the decoding. For example, the model might output synonyms that have a high embedding similarity to words in the reference text. We propose several alternative decoding algorithms to better achieve the goal.
% and compare them against to the widely used beam search algorithm in the experiments.

\subsubsection{Unconstrained Decoding}
The original decoding target is to find $\hat{\bm y}$ that satisfies
\begin{equation}
\begin{split}
    \hat{\bm y} &= {\arg\max}_{\bm y}\log P(\bm y|\bm{\mathcal{Z}};\bm \theta_\text{EC})
\end{split}
\end{equation}
where $\bm{\mathcal{Z}} = \{\bm{\hat z}^{(1)}, \bm{\hat z}^{(2)}, \cdots, \bm{\hat z}^{(n)}\}$ is the input to the N-best T5 encoder containing $n$ ASR hypotheses. Since it is time-consuming to find the globally optimal sequence satisfying this equation, heuristic algorithms such as beam search are often used in decoding. Since no explicit constraints about the generated sequence are applied during decoding, we refer to this method as unconstrained decoding in this paper.

\subsubsection{N-best Constrained Decoding}
The original decoding space of the error correction model is unbounded. However, we expect the generated correction results to sound similar to the original utterance. The N-best list contains hypotheses generated by the ASR system which are most likely to be the correct transcription. In N-best constrained decoding, we force the generation results to sentences within this ASR N-best list. Furthermore, since each path in the N-best list is also associated with a score calculated by the ASR system, we can combine the scores from two models with an interpolation weight $\lambda$. To be specific, $\hat{\bm y}$ is the decoding result that maximizes Equation \ref{eq:n-best}. $\bm x$ and $\bm{\mathcal{Z}}$ refer to the input acoustic features of the ASR system and the obtained ASR N-best list.
\begin{equation}
\begin{split}
    \hat{\bm y} &= {\arg\max}_{\bm y\in \bm{\mathcal{Z}}} [(1-\lambda) \cdot \log P(\bm y|\bm{x};\bm\theta_\text{ASR})\\
    &+ \lambda \cdot \log P(\bm y|\bm{\mathcal{Z}};\bm \theta_\text{EC})]
\end{split}
\label{eq:n-best}
\end{equation}

% When $\lambda$ is set to $\bm 1$, we rescore the $n$-best list only based on the probabilities calculated by the text correction model. Otherwise, we combine the ASR scores in the decoding process.

\subsubsection{Lattice Constrained Decoding}
An ASR N-best list can only encode a small subset of all the possible decoding results. Therefore, constraining the decoding result to appear in the N-best list might be a strict condition. Another data structure to consider is the lattice generated with path merging. Here, we restrict the decoding space to paths $\bm{\mathcal{G}}$ in the lattice and combine ASR scores in the decoding. 

\begin{equation}
\begin{split}
    \hat{\bm y} &= {\arg\max}_{\bm y\in \bm{\mathcal{G}}} [(1-\lambda) \cdot \log P(\bm y|\bm{x};\bm\theta_\text{ASR})\\
    &+ \lambda \cdot \log P(\bm y|\bm{\mathcal{Z}};\bm \theta_\text{EC})]
\end{split}
\end{equation}

% When path $\bm y$ does not exist in lattice graph $\mathcal{G}$, probability $P_\mathcal{G}(\bm y_{1:N})$ is equal to $\bm 0$.

\begin{figure}[!htbp]
\centering
\includegraphics[width=0.27\textwidth]{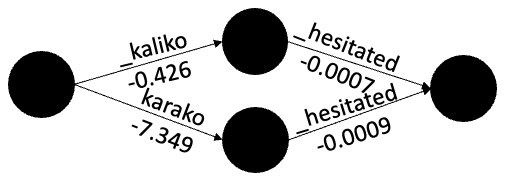}
\caption{Example lattice with words.}
\label{word}
\vspace{-3mm}
\end{figure}

\begin{figure}[!htbp]
\centering
\includegraphics[width=0.48\textwidth]{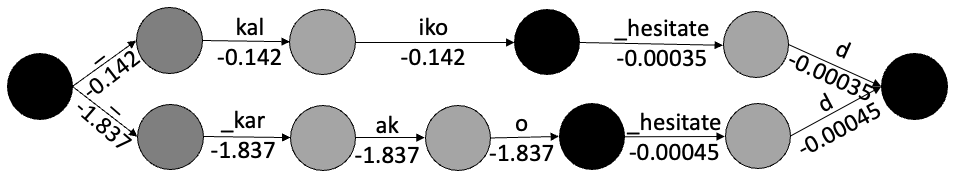}
\caption{Example lattice with PLM BPE tokens.}
\label{lm_bpe}
\vspace{-3mm}
\end{figure}

Since the ASR model and the pre-trained language model use different tokenizers, we need to convert the original lattice into an equivalent form suitable for the N-best T5 to process. The lattice with ASR BPE tokens is first converted into a word lattice with dynamic programming as in Figure \ref{word}. The words on the edges are then split into BPE tokens by the tokenizer of T5, which is shown in Figure \ref{lm_bpe}. The detailed decoding algorithm is shown in Algorithm \ref{lattice_decode}, which is adapted from \cite{auli2013joint}. A similar algorithm is proposed in \cite{ma2020neural} while the lattice is obtained from different generation processes.

% TODO: beam width k???

\begin{algorithm}[!htbp]
\footnotesize
\caption{Lattice Constrained Decoding for N-best T5}
\label{lattice_decode}
\textbf{Data:} lattice node set $\mathcal{V}$, lattice edge set $\mathcal{E}$, beam width $b$, T5 encoder outputs $\{\textbf h_j\}$

\begin{algorithmic}[1]
\STATE $Q\leftarrow \text{topological\_sort}(\mathcal{V})$
\FOR{$v$ in $\mathcal{V}$}
\STATE $H_v\leftarrow$ min\_heap()
\ENDFOR
\STATE $n_0.\text{history}=\epsilon, n_0.\text{score}=0$
\STATE $H_\text{start}$.put($n_0$) %\COMMENTA{Initialize start-state}
\FOR{$v$ in $Q$}
\STATE $\mathbf o=\text{Decoder}(\{\textbf h_j\}, n.\text{history}, v.\text{word}$)
\FOR{$\langle v,x\rangle$ in $\mathcal{E}$}
\FOR{$n$ in $H_v$}
\STATE $n^\prime.\text{history}=\text{concat}(n.\text{history},v.\text{word})$
\STATE $n^\prime.\text{score}=n.\text{score}+\lambda\cdot \log(\mathbf o[x.\text{word}])+(1-\lambda)\cdot\log s_{vx}$
\vspace{-3mm}
\IF{$H_x.\text{size}\ge b\land H_x.\text{score}.\text{min}()<n^\prime.\text{score}$}
\STATE $H_x$.remove\_min()
\ENDIF
\IF{$H_x.\text{size}< b$}
\STATE $H_x$.put($n^\prime$)
\ENDIF
\ENDFOR
\ENDFOR
\ENDFOR
\STATE $I=$ max\_heap($H_\text{end}$.items)
\RETURN $I$.max()
\end{algorithmic}
% \vspace{-5mm}
\end{algorithm}
% \vspace{-5mm}

\section{Experiments}
\subsection{Experiment Setup}
We conduct experiments on the LibriSpeech dataset \cite{panayotov2015librispeech}. The data is collected from audiobook readings and the training set contains 960hr of speech data. We test on four subsets: dev\_clean, dev\_other, test\_clean, and test\_other, each containing $\sim$5hr of speech. 80-dimensional log-mel filterbank features are extracted from the utterances and the sample rate is 16kHz.

The ASR model utilized in this paper adopts a novel Conformer-Transducer structure \cite{gulati2020conformer}. The encoder contains 12 Conformer layers with a hidden size of 512. The predictor has one LSTM layer. The hidden dimensions of the jointer and predictor are both 512. SpecAugment \cite{park2019specaugment} and speed perturbation are used for training data augmentation. The hyper-parameters of this Transducer model follow the ESPnet code base \cite{watanabe2018espnet}.

% SpecAugment with two frequency masks with mask parameter $F=30$, 2 time masks with mask parameter $T=40$, and time warping with warp parameter $W=5$ is used during training. 

% \subsection{Error Correction Model Setup}
The error correction model learns from the erroneous transcriptions generated by the ASR model. Since the ASR model is trained on the training speech data, it shows an extremely low WER rate (close to zero) on this set. Data augmentation techniques are needed so that the error correction model generalizes to speech data unseen by the ASR model. Past works have used different data generation methods including decoding with under-performed ASR model \cite{zhao2021bart}, usage of dropout in the ASR decoding \cite{hrinchuk2020correction}, and decoding on pseudo TTS data \cite{guo2019spelling}, etc. 

% To generate training data for the error correction model, we need to conduct perturbation on the original ASR training data. 

In this work, we only use one simple yet effective data augmentation method, SpecAugment. Two frequency masks with mask parameter $F=30$, eight time masks with mask parameter $T=40$, and time warping with warp parameter $W=40$ are utilized on the training speech data. We decode the ASR model on this perturbed data set and filter out sentences with WER higher than 0.25. The resulting training text corpus contains 262K sentence pairs. It is worth noting that in this work we do not leverage the extra text corpus of LibriSpeech. We assume that only the training speech is available for the error correction task, which is a more general case. Therefore, our training pipeline can be easily adapted to any speech corpus.

The error correction model is initialized using a pre-trained T5 base model. The model has 6 Transformer blocks for both the encoder and decoder with a hidden dimension of 768. We train the model on the perturbed training data for 3 epochs. AdamW \cite{loshchilovdecoupled} is utilized as the optimizer and the initial learning rate is 5e-5. In fine-tuning, the dropout rate is set to 0.1. A batch size of 32 is used for both training and testing. In the constrained decoding process, optimal interpolation weight $\lambda$ is searched in the range of [0.0, 1.0] with a grid size of 0.05. In lattice constrained decoding, the beam width $b$ is set to 1, and larger values yield similar results.

% \subsection{Analysis on ASR Output}
\subsection{Error Correction Results}
\begin{table}[!htbp]
    \centering
    \footnotesize
    \caption{ASR system performance.}
    \vspace{-2mm}
    \begin{tabular}{l|cc|cc}
    \toprule
        \multirow{2}*{Data} & \multicolumn{2}{c|}{Dev} & \multicolumn{2}{c}{Test} \\
        & clean & other & clean & other \\
        \midrule
        Beam Search & 2.71 & 6.99 & 2.88 & 7.06 \\
        % Path Merging $(K=3)$ \\
        + Path Merging & 2.70 & 7.01 & 2.91 & 7.13 \\
        % Path Merging $(K=5)$ \\
        \bottomrule
    \end{tabular}
    \label{tab:asr}
    \vspace{-2mm}
\end{table}

In Table \ref{tab:asr} we show the baseline results of the ASR model. Here we list the WER results of the 1-best decoding hypothesis obtained with and without path merging in the beam search. For both methods, a beam width of 10 is used in the decoding. Therefore, an N-best list containing the 10 best hypotheses is generated during decoding. For path merging, similar paths are merged with a context size of $k=4$. The results are similar to~\cite{prabhavalkar2021less} in that the performance of the ASR system decoded with the path-merging scheme shows comparable performance to the counterpart using full-context history.

\begin{table}[!htbp]
    \centering
    \footnotesize
    \caption{Oracle WER results in ASR outputs.}
    % \vspace{-2.5mm}
    \begin{tabular}{l|cc|cc}
    \toprule
        \multirow{2}*{ASR Output} & \multicolumn{2}{c|}{Dev} & \multicolumn{2}{c}{Test} \\
        & clean & other & clean & other \\
        % \midrule
        % Baseline & 2.71 & 6.99 & 2.88 & 7.06 \\
        \midrule
        5-best List & 1.35 & 4.72 & 1.44 & 4.66 \\
        10-best List & 1.24 & 4.43 & 1.34 & 4.34 \\
        % \midrule
        % Path Merging 10-best & 2.40 & 6.23 & 2.56 & 6.35 \\
        % Lattice 10-best & 1.13 & 4.14 & 1.26 & 4.08 \\
        % Lattice 50-best & & & 0.97 & 3.26 \\
        Lattice & 0.79 & 2.98 & 0.89 & 3.00 \\
        \bottomrule
    \end{tabular}
    \label{tab:oracle}
    % \vspace{-5mm}
\end{table}

Table \ref{tab:oracle} shows the oracle WER results in different ASR outputs. The 5-best list contains the top 5 hypotheses from the 10-best list. The oracle WER in the 5-best list and 10-best list improve by 38.6\% and 42.8\% compared to the ASR baseline. Therefore, words within the N-best list can help the model to recover the correct transcription. By merging partial hypotheses with similar sentence history, we can generate word lattices in the decoding. As shown in the table, the lattice yields lower oracle WER results on the test sets. Results indicate that N-best lists and lattices can provide guidance in the decoder of the error correction model to achieve controllable decoding.
% Since the ground-truth transcription is likely to appear in the lattice graph, which is useful for providing guidance in the lattice-constrained decoding algorithm. 

% \subsection{N-best T5 Training}
\begin{table}[!htbp]
    \centering
    \footnotesize
    \caption{Results for ASR baseline and N-best T5 models with different model inputs.}
    % \vspace{-2.5mm}
    \begin{tabular}{l|cc|cc}
    \toprule
    \multirow{2}*{Model} & \multicolumn{2}{c|}{Dev} & \multicolumn{2}{c}{Test} \\
    & clean & other & clean & other \\
    \midrule
    Baseline & 2.71 & 6.99 & 2.88 & 7.06 \\
    \midrule
    1-best T5 & 2.89 & 6.94 & 3.02 & 7.18 \\
    5-best T5 & 2.62 & 6.40 & 2.77 & 6.67 \\
    10-best T5 & \textbf{2.60} & \textbf{6.25} & \textbf{2.67} & \textbf{6.56} \\
    \bottomrule
    \end{tabular}
    \label{tab:t5_result}
    % \vspace{-5mm}
\end{table}

The results of training an error correction model with different model inputs are listed in Table \ref{tab:t5_result}. In the decoding of the error correction model, the beam search algorithm with a beam size of 10 is used.
% For the proposed N-best T5 model, we testify two different kinds of training targets. ``Ref'' is short for ``Reference'' and refers to the standard way of using the reference text in the training. ``Ora'' stands for ``Oracle'', which means we use the sentence that yields the lowest WER in the N-best list as the training target. 
From the table, we can see that the model fails to yield good results only using the 1-best decoding hypothesis as model input. Possible reasons have been discussed in this paper that without enough input information, the model fails to improve performance over a strong ASR system. For the 5-best T5 and the 10-best T5 model, 4.7\% and 7.2\% performance gains could be seen compared to the baseline ASR model. The results indicate that with a larger N, the error correction model obtains richer information from diversified input, which leads to more accurate error detection and correction.

\subsection{Comparison of Decoding Algorithms}

In this experiment, we compare the WER results of the N-best T5 model using different decoding algorithms. As shown in Table \ref{tab:t5_rescore}, by adding more constraints in the decoding process, the model performance is improved for both 5-best T5 and 10-best T5 models. With the proposed methods, we can guide the error correction model to generate homophones of the mistaken words. Since a lattice contains more possible paths compared to an N-best list and yields lower oracle WER, lattice-constrained decoding gives slightly better performance than N-best constrained decoding on the test sets.

\begin{table}[!htbp]
    \centering
    \footnotesize
    \caption{Comparison of ASR baseline and N-best T5 model with different decoding algorithms.}
    \vspace{-2.5mm}
    \begin{tabular}{l|l|cc|cc}
    \toprule
        \multirow{2}*{Model} & \multirow{2}*{Decoding Method} & \multicolumn{2}{c|}{Dev} & \multicolumn{2}{c}{Test} \\
        && clean & other & clean & other \\
        \midrule
        Baseline & - & 2.71 & 6.99 & 2.88 & 7.06 \\
        \midrule
        % \multirow{3}*{1-best T5} & unconstrained & - & 2.89 (6.6\%) & 6.94(-0.8\%) & 3.02 (4.9\%) & 7.18 (1.7\%) \\
        % & 10-best Rescore & 1.0 &&& 2.82(-2.08\%) & 6.94(-1.70\%) \\
        % & Lattice Rescore \\
        % \hline
        % \multirow{3}*{5-best T5} & unconstrained & - & 2.62(-3.32\%) & 6.40(-8.44\%) & 2.77(-3.82\%) & 6.67(-5.52\%) \\
        % & 5-best Rescore & 0.75 & 2.39(-11.8\%) & 6.27(-10.3\%) & 2.54(-11.8\%) & 6.39(-9.49\%) \\
        % & Lattice Rescore &&&& 2.55 & 6.34 \\
        % % & 10-best Rescore &  \\
        % \hline
        \multirow{3}*{5-best T5} & Unconstrained & 2.62 & 6.40 & 2.77 & 6.67 \\
        & N-best Constrained & \textbf{2.38} & 6.25 & 2.55 & 6.38 \\
        % & Lattice 10-best & \\
        & Lattice Constrained & 2.40 & 6.21 & 2.54 & 6.34 \\
        % & Lattice 10-best & 2.41 & 6.28 & 2.58 & 6.40 \\
        % & Lattice & 2.41 & 6.26 & 2.58 & 6.39 \\
        \midrule
        \multirow{3}*{10-best T5} & Unconstrained & 2.60 & 6.25 & 2.67 & 6.56 \\
        & N-best Constrained & 2.39 & 6.17 & 2.54 & 6.31 \\
        % & Lattice 10-best & 2.40 & 6.18 & 2.54 & 6.31 \\
        & Lattice Constrained & 2.41 & \textbf{6.11} & \textbf{2.53} & \textbf{6.27} \\
        % & Lattice 10-best & 2.42 & 6.06 & 2.53 & 6.22 \\
        % % && Lattice 50-best & 0.75 &&& 2.53 & 6.27 \\
        % & Lattice & 2.42 & \textbf{6.00} & \textbf{2.52} & \textbf{6.16}\\
        \bottomrule
    \end{tabular}
    \label{tab:t5_rescore}
    % \vspace{-5mm}
\end{table}

\begin{figure}[ht]
\centering
\includegraphics[width=0.4\textwidth]{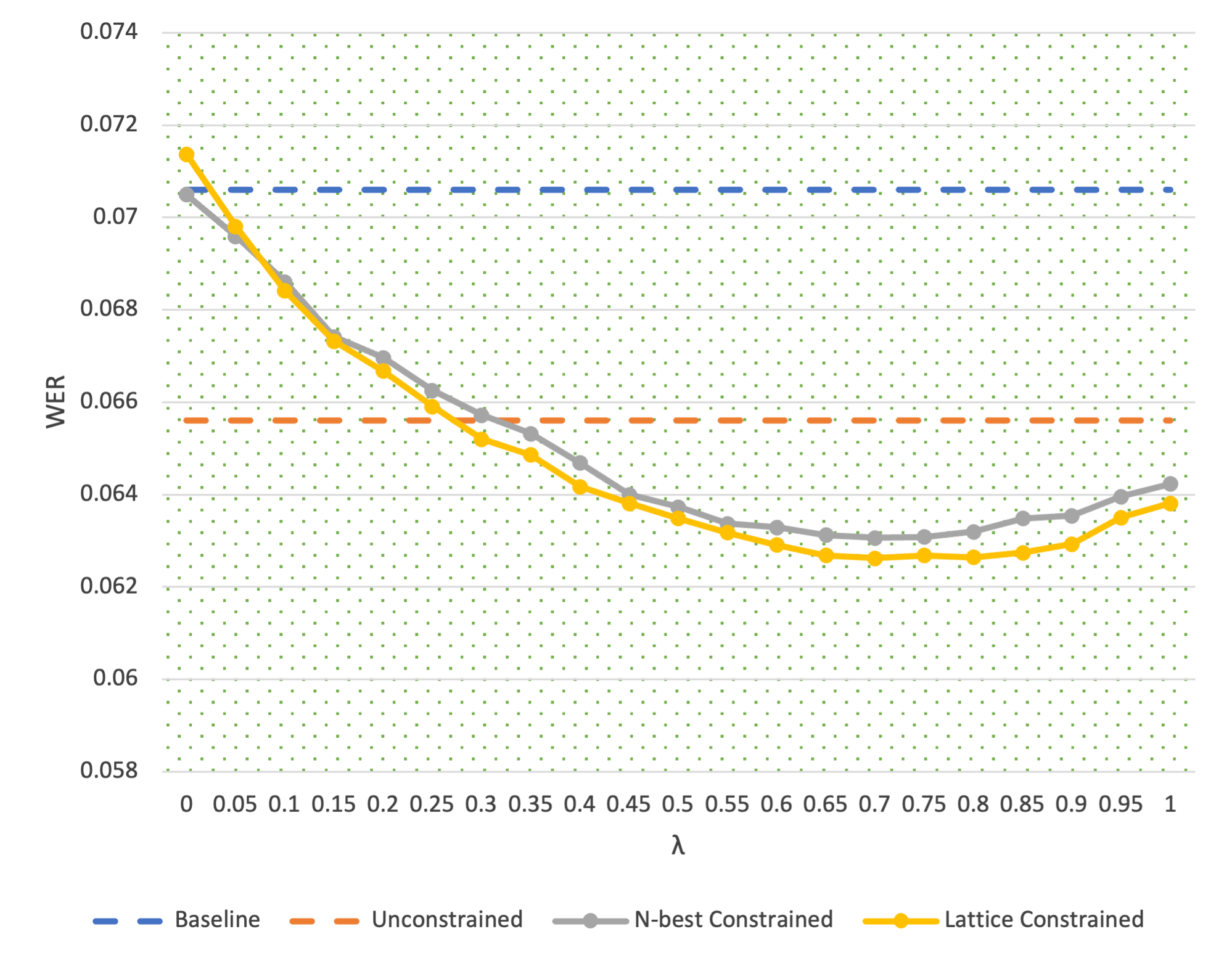}
\vspace{-2mm}
\caption{Performance of 10-best T5 error correction model on test\_other with different interpolation weights $\lambda$.}
\label{lambda}
\vspace{-5mm}
\end{figure}

The effect of interpolation weight $\lambda$ on the constrained decoding is depicted in Figure \ref{lambda}. $\lambda$ is the weight of the error correction model. As the figure shows, constrained decoding with $\lambda=1$ outperforms the plain beam search. In these cases, we constrain the error correction model to only generate hypotheses that are in the ASR N-best list or lattice rather than the entire decoding space. When $0<\lambda<1$, the scores calculated by the ASR system are linearly combined to bias the result towards paths with higher acoustic probabilities. The best performance is achieved when $\lambda$ is around 0.75.

\section{Conclusions}
In this work, we propose an effective error correction model based on transfer learning from pre-trained language models. The performance of the proposed method consistently improves over the performance of a strong ASR model on the LibriSpeech test sets. The two contributions of this paper: the usage of the N-best list as input to PLMs and constrained decoding algorithms based on the output from E2E ASR models can be applied to other models within the error correction area.

% \section{Acknowledgements}

\bibliographystyle{IEEEtran}
\bibliography{mybib}

\end{document}